\documentclass[runningheads]{llncs}

\usepackage{eccv}

\usepackage{eccvabbrv}

\usepackage{graphicx}
\usepackage{booktabs}

\usepackage[accsupp]{axessibility}  

\usepackage{hyperref}

\usepackage{orcidlink}

\def\eg{\emph{e.g.}} 

\def\ie{\emph{i.e.}}

\def\vs{\emph{vs. }}
\def\al{\emph{et al. }}
\def\wrt{w.r.t. }

\usepackage{url}

\usepackage{amsthm}
\usepackage{booktabs}
\usepackage{algorithm}
\usepackage{algorithmic}
\usepackage{array}
\usepackage{multirow}
\usepackage{graphicx}
\usepackage{xcolor,colortbl}
\usepackage{color}
\usepackage{colortbl}
\usepackage[misc]{ifsym}
                           
\usepackage{float} 
\usepackage{wrapfig}

\usepackage{bbding}

\begin{document}

\title{Foster Adaptivity and Balance in Learning with Noisy Labels} 

\titlerunning{Foster Adaptivity and Balance in Learning with Noisy Labels}

\author{Mengmeng Sheng\inst{1}\orcidlink{0000-0002-2011-8597} \and
Zeren Sun\inst{1}\textsuperscript{(\Letter)}\orcidlink{0000-0001-6262-5338} \and
Tao Chen\inst{1}\orcidlink{0000-0001-8239-1698} \and
Shuchao Pang\inst{1}\orcidlink{0000-0002-5668-833X} \and
Yucheng Wang\inst{2}\orcidlink{0000-0002-8290-3291} \and
Yazhou Yao\inst{1}\textsuperscript{(\Letter)}\orcidlink{0000-0002-0337-9410}}

\authorrunning{M. Sheng et al.}

\institute{Nanjing University of Science and Technology, Nanjing, China\\ 
\email{\{shengmengmemg, zerens, taochen, pangshuchao, yazhou.yao\}@njust.edu.cn} \and
Horizon Robotics, Beijing, China\\
\email{yucheng.wang@horizon.cc}}

\maketitle

\begin{abstract}
Label noise is ubiquitous in real-world scenarios, posing a practical challenge to supervised models due to its effect in hurting the generalization performance of deep neural networks.
Existing methods primarily employ the sample selection paradigm and usually rely on dataset-dependent prior knowledge (\eg, a pre-defined threshold) to cope with label noise, inevitably degrading the adaptivity. Moreover, existing methods tend to neglect the class balance in selecting samples, leading to biased model performance.
To this end, we propose a simple yet effective approach named \textbf{SED} to deal with label noise in a \textbf{S}elf-adaptiv\textbf{E} and class-balance\textbf{D} manner. 
Specifically, we first design a novel sample selection strategy to empower self-adaptivity and class balance when identifying clean and noisy data.
A mean-teacher model is then employed to correct labels of noisy samples.
Subsequently, we propose a self-adaptive and class-balanced sample re-weighting mechanism to assign different weights to detected noisy samples.
Finally, we additionally employ consistency regularization on selected clean samples to improve model generalization performance.
Extensive experimental results on synthetic and real-world datasets demonstrate the effectiveness and superiority of our proposed method.
The source code has been made anonymously available at \textcolor[HTML]{da3593}{\url{https://github.com/NUST-Machine-Intelligence-Laboratory/SED}}.


\end{abstract}

\section{Introduction}
Deep neural networks (DNNs) have witnessed remarkable achievements in many computer vision tasks, such as image classification \cite{Mao,Jiang}, object detection \cite{YOLO,Faster_RCNN}, face recognition \cite{Face_Recognition}, and instance segmentation \cite{Chen_1,Chen_2}.
The superior performance of DNNs is highly attributed to supervised training with large-scale and high-quality human-labeled training datasets (\eg, ImageNet \cite{InageNet}).
However, collecting large-scale datasets with accurate annotations is expensive and time-consuming, especially for tasks requiring expert annotation knowledge (\eg, medical images \cite{Medical_Images}). 
To alleviate this problem, researchers start to resort to alternative methods, such as crowd-sourcing platforms \cite{Crowd-sourcing} or web image search engines \cite{Sesrch_engine}, for obtaining cheaper label annotations.
Unfortunately, these methods usually result in unavoidable noisy labels, which tend to cause inferior model performance due to the strong learning ability of DNNs \cite{zhang2016understanding}.
Consequently, developing robust models for learning with noisy labels is of significant importance.

\begin{figure}[t]
\centering
\includegraphics[width=\linewidth]{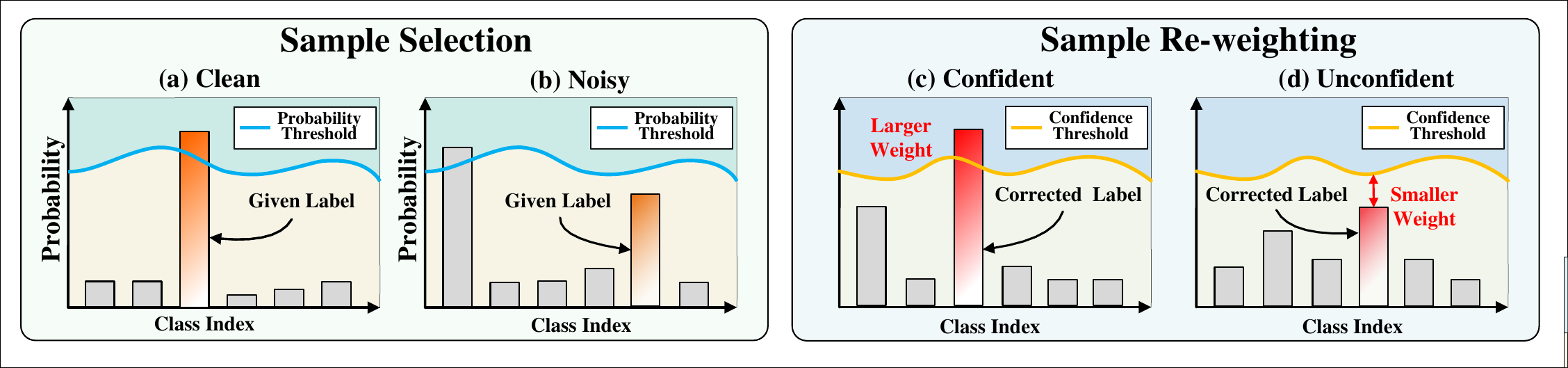}
\vspace{-0.5cm}
\caption{(a-b) Self-adaptive and class-balanced sample selection based on predicted probability \wrt given labels. The blue curve indicates the class-specific selection thresholds. (c-d) Self-adaptive and class-balanced sample re-weighting based on correction confidence. The orange curve represents the class-specific confidence threshold.}
\label{figure1}
\vspace{-0.5cm}
\end{figure}

Recently, a growing number of methods have been proposed for addressing the label noise problem \cite{LiuTongliang_v1_NIPS, LiuTongLiang_v2_TPAMI, LiuTongLiang_v4_CVPR, LiuTongLiang_v5_CVPR, LiuTongLiang_v6_ICML, LiuTongLiang_v7_ICML, HANBO_v1_TPAMI, SANM, CVPR23}.
Label correction and sample selection/re-weighting are two major strategies for tackling noisy labels.
Label correction methods typically attempt to rectify labels using the noise transition matrix \cite{goldberger2017} or model predictions \cite{dividemix}. 
For example, \cite{patrini2017making} proposes to correct corrupted labels by estimating the noise transition matrix.
Jo-SRC \cite{josrc} uses the temporally averaged model (\ie, mean-teacher model) to generate reliable pseudo-label distributions for providing supervision.
However, on the one hand, the noise transition matrix is hard to estimate in real-world scenarios. 
On the other hand, networks tend to have better recognition capability on simple categories than hard ones.
This recognition bias usually results in imbalanced label corrections (\ie, samples are more likely to be corrected into simple categories) in prediction-based label correction methods, hurting the final model performance.

Another line of research focuses on the sample selection/re-weighting \cite{co-teaching,yao2021dual, josrc, MentorNet, NPN, CBS,sun2020crssc,sun2021webly}.
Sample selection methods primarily seek to split samples into two subsets: a noisy subset and a clean subset \cite{co-teaching, josrc, Small_Loss}.
Prior methods tend to regard samples with small losses as clean ones \cite{co-teaching, JoCoR}.
For example, JoCoR \cite{JoCoR} exploits a joint loss to select small-loss samples to encourage agreement between models.
However, these methods often require proper prior knowledge (\eg, a pre-defined drop rate or threshold) to achieve effective sample selection.
Moreover, previous literature usually neglects class balance during sample selection, leading to biased model performance.
Sample re-weighting can be deemed as a variant of sample selection, smoothing its 0/1 weighting scheme to a softer one. 
Samples with higher confidence are assigned larger weights, while those with lower confidence are assigned smaller weights.
For example, L2RW \cite{L2RW} proposes to assign different sample weights based on meta-learning.
However, existing sample re-weighting methods also tend to require prior knowledge (\eg, a small subset of clean samples).

To alleviate the aforementioned issues, we propose a simple yet effective method, named \textbf{SED}, to learn with noisy labels in a \textbf{S}elf-adaptiv\textbf{E} and class-balance\textbf{D} manner.
Our SED integrates sample selection, label correction, and sample re-weighting.
Specifically, we propose to identify clean samples based on the predicted probability \wrt the given labels of input samples.
To promote self-adaptivity and class balance in sample selection, we propose to integrate global and local thresholds for each category when distinguishing between clean and noisy data (as shown in Fig.~\ref{figure1} (a) and (b)). 
The global and local thresholds are dynamically updated during training.
Once the clean and noisy subsets are obtained, we employ a mean-teacher model to correct labels for identified noisy samples.
Subsequently, we propose to re-weight label-corrected noisy samples in a self-adaptive and class-balanced fashion to alleviate the confirmation bias caused by imbalanced label correction. 
We impose larger/smaller weights on noisy samples with higher/lower correction confidence according to an estimated truncated normal distribution (as shown in Fig.~\ref{figure1} (c) and (d)).
Finally, we employ an additional regularization loss term on identified clean samples to further enhance the performance and robustness of the model.
Comprehensive experimental results have been provided to verify the effectiveness and superiority of our proposed SED on synthetically corrupted datasets and real-world datasets.
Our contributions are summarized as follows: 

(1) We propose a simple yet effective method, named SED, to combat noisy labels. 
SED selects and re-weights samples in a self-adaptive and class-balanced manner, alleviating the demand for dataset-dependent prior knowledge and the negative effect caused by class imbalance.

(2) Our proposed SED selects samples according to class-specific thresholds that are estimated in a data-driven manner, encouraging self-adaptivity and class balance in sample selection.
In addition, we propose to re-weight samples based on a truncated normal distribution that is updated periodically, mitigating performance downgrade due to imbalanced label corrections.

(3) We provide comprehensive experimental results on synthetic and real-world datasets to illustrate the superiority of our proposed SED. 
Extensive ablation studies are conducted to further verify the effectiveness of our method.

\begin{figure*}[t]

\centering
\includegraphics[width=\linewidth]{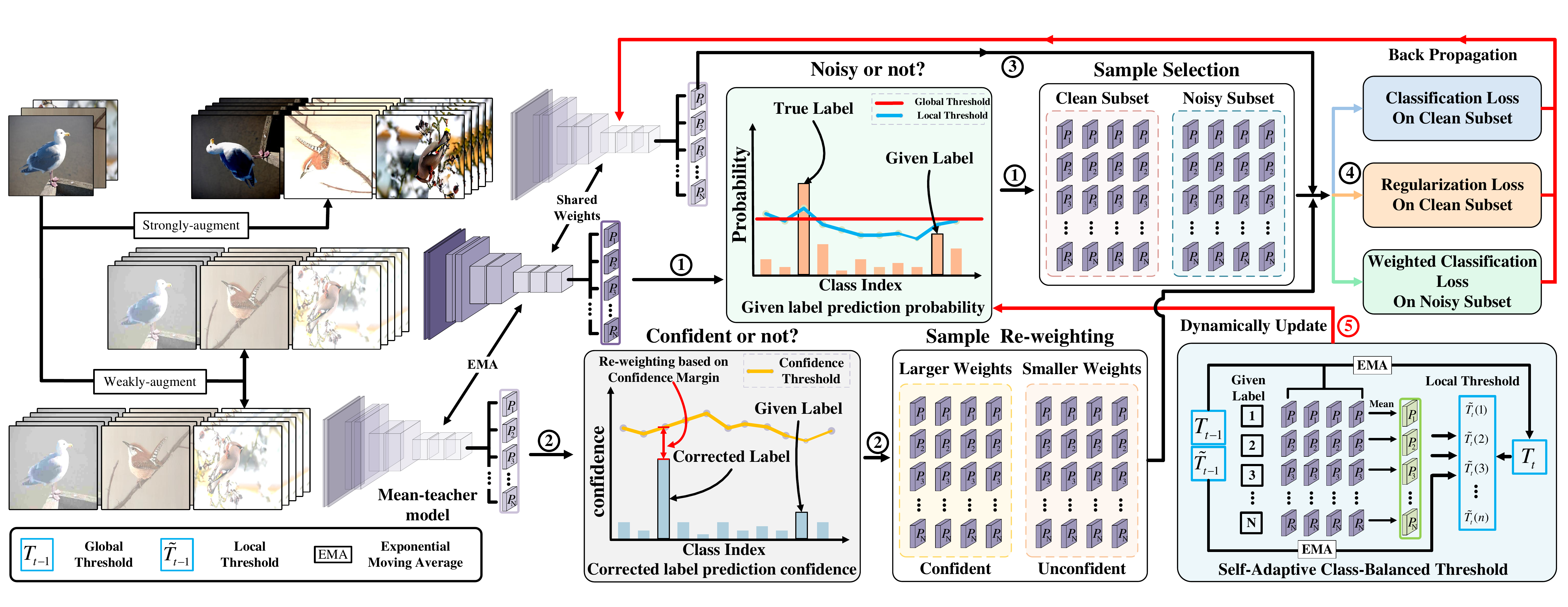}
\vspace{-0.4cm}
\caption{The overall framework of our SED. We first divide the training set into a clean subset and a noisy subset based on global and local thresholds that are dynamically updated. Our threshold design enables self-adaptivity and class balance in sample selection. We then employ a mean-teacher model to correct labels for noisy samples. Based on the correction confidence, SED adaptively assigns different weights to label-corrected noisy samples and uses them for training. Finally, SED further boosts the model performance by imposing an additional consistency regularization loss on selected clean samples. The final objective loss integrates the classification losses on clean and noisy samples and the regularization loss on clean samples.}
\label{figure2}
\vspace{-0.5cm}
\end{figure*}
\section{Related Work} 

\noindent
\textbf{Label Correction.}
The intuitive idea for handling noisy labels is to correct corrupted labels before feeding them into networks \cite{Goldberger2016TrainingDN, patrini2017making, xia2019anchor,yao2021dual,goldberger2017,HANBO_v2_CVPR,Liu_1,Liu_2,ProSelfLC}.
Early works propose to correct the training labels by estimating the noise transition matrix.
\cite{yao2021dual} introduces an intermediate class to avoid directly estimating the noisy class posterior and then factorizes the transition matrix into the product of two sub-matrices.
However, the transition matrix is hard to estimate accurately in real-world scenarios.
Some other methods propose to model label noise by using predictions of DNNs \cite{PENCIL,lee2018cleannet,vahdat2017robustness,veit2017learning}.
\cite{PENCIL} proposes to directly learn label distributions for corrupted samples in an end-to-end manner. 
Nevertheless, since DNNs tend to learn better on simple categories than hard ones, pseudo-labels are more likely to fall into the simple class set, leading to imbalanced label correction.
In this work, we resort to the re-weighting strategy to alleviate the issue caused by imbalanced label correction.

\noindent
\textbf{Sample Selection.} 
Another type of classical method to deal with label noise is sample selection, which divides the training set into a clean subset and a noisy subset \cite{co-teaching, josrc, sun2022pnp, JoCoR}. 
Previous sample selection methods primarily employ the cross-entropy loss as the selection criterion, regarding samples with small losses as clean ones.
For example, Co-teaching \cite{co-teaching} proposes to cross-update two networks using small-loss samples selected by peer networks.
Some recent methods propose new selection criteria for finding clean samples \cite{josrc, dividemix}.
Jo-SRC \cite{josrc} proposes to employ Jensen-Shannon Divergence for selecting clean samples globally.
DISC \cite{DISC} proposes to select reliable instances based on the insight of memorization strength.
However, these methods usually demand pre-defined drop rates or thresholds.
Furthermore, previous methods neglect the class imbalance issue in the selection process, leading to inferior and biased model performance.
In this work, we employ predicted probability as the selection criterion and propose a novel threshold mechanism to enable self-adaptive and class-balanced selection. 

\noindent
\textbf{Sample Re-weighting.}
Recently, some researchers have been devoted to re-weighting training samples to cope with noisy labels \cite{Meta-Weight-Net, HANBO_v3_ICLR, ReWeighting1, DMLP}.
These methods usually assign larger weights to samples that are more likely to be clean while smaller weights to others, minimizing the misleading impact of noisy samples.
For example, L2RW \cite{L2RW} proposes a meta-learning algorithm that learns to assign weights to training examples based on their gradient directions. 
However, existing methods tend to require considerable prior knowledge (\eg, a small subset of clean samples), posing a limit to their practicability.
In this work, we design a novel re-weighting scheme to empower self-adaptivity and class balance when leveraging label-corrected noisy samples.

\section{Method}
\subsection{Problem Statement}
Formally, considering a $C$-class classification problem, we denote $D_{train}=\{(x_i, y_i)|i=1,...,N \}$ as the training set with label noise, in which $x_i$ denotes the $i$-th training sample and $y_i \in \{0,1\}^C$ is its associated label (potentially ``incorrect''). 
We use $y_i^*$ to represent the ground-truth label of $x_i$ and denote  $D_{test}=\{(x_i, y_i^*)|i=1,...,M \}$ as the test set with accurate labels.
$N$ and $M$ represent the total number of samples in the training set $D_{train}$ and test set $D_{test}$, respectively. 
The goal is to train a robust classification neural network $\mathcal{F}(\cdot,\theta)$ ($\theta$ denotes network parameters) on the noisy training set $D_{train}$ to perform accurate prediction on the test set $D_{test}$.
The conventional classification task usually hypothesizes that given labels of training samples are accurate (\ie, $y_i = y_i^*$), thus using the following cross-entropy loss to optimize the network.
\begin{equation}\label{eq_ce}
\mathcal{L}_{ce} = -\frac{1}{N}\sum_{i=1}^{N}\sum_{c=1}^{C}y_i^clog(p^c(x_i, \theta)),
\end{equation}
in which $p^c(x_i, \theta)$ denotes the predicted softmax probability of the $i$-th training sample $x_i$ over its $c$-th class.

Due to the memorization effect \cite{memorization-effect} (\ie, models tend to fit clean and simple samples first and then gradually memorize noisy ones), the network optimization based on the above loss usually leads to an ill-suited solution. 
One potentially useful remedy is to integrate sample selection, label correction, and sample re-weighting.
In this work, we follow this paradigm to combat noisy labels by encouraging self-adaptivity and class balance.

\subsection{Adaptive and Balanced Sample Selection}\label{SCS}
Previous studies \cite{co-teaching, JoCoR, josrc} usually select small-loss samples as clean ones based on pre-defined drop rates or thresholds.
It should be noted that drop rates can be easily converted to thresholds during selection, thus we only discuss thresholds hereafter.
The selection thresholds are usually dataset-dependent, making it challenging to adapt them to different real-world datasets.
Although existing methods employ scheduling strategies (\eg, a gradually increasing schedule \cite{co-teaching}) to adjust thresholds during training for fully exploiting the model capability, these scheduling designs are rather heuristic and still require pre-defined initial and final threshold values.
Moreover, few works consider the different difficulties in learning various categories, leading to biased selection results and inferior model performance. 

To this end, we propose a self-adaptive and class-balanced sample selection (SCS) strategy to address the above problems.
SCS adaptively adjusts the threshold in an epoch-wise and class-wise manner to enable effective clean sample identification.
Specifically, we employ global and local thresholds, which are both self-adaptive, to distinguish between clean and noisy samples in each category.
Since the cross-entropy loss is unbounded, we propose to rely on predicted probability \wrt the given labels $p^{y_i}(x_i, \theta)$ to determine whether the samples are clean. 
Samples with higher $p^{y_i}(x_i, \theta)$ are more likely to have correct labels.

We estimate the global threshold based on the averaged predicted probability \wrt given labels over all training samples to reflect the overall learning state of the network. 
This design makes the global threshold data-driven, thus eliminating the demand for pre-defined thresholds.
Moreover, we employ the exponential moving average (EMA) to further refine the global threshold, alleviating unstable training caused by large perturbation of the averaged predicted probability. By adopting an initial value of $T_0 = \frac{1}{C}$, our final global threshold at the $t$-th epoch is defined as:
\begin{equation}\label{eq:3}
T_t = \left\{
	\begin{array}{ll}
		\frac{1}{C}, & t=0 \\
        m T_{t-1} + (1-m) \frac{1}{N} \sum_{i=1}^N p^{y_i}(x_i, \theta), & t > 0
	\end{array}.
\right.
\end{equation}
Our design of the global threshold scheduling implicitly complies with the memorization effect \cite{memorization-effect}. 
As the training progresses, the predicted probability \wrt the given label gradually increases, leading to the monotonic increase of $T_t$.
Consequently, the network can learn from more samples in the early stage but fewer samples in the later stage.

As stated above, using only a global threshold to divide the training set neglects the difference among various categories and will result in imbalanced sample selection (\ie, fewer samples of complicated categories will be selected as clean data). 
Samples of easy categories tend to be better learned and have higher $p^{y_i}(x_i, \theta)$, thus requiring larger thresholds to distinguish between clean and noisy data.
Therefore, we additionally propose a local threshold scheme to further adjust the global threshold.
We first estimate the expectation of the model’s predictions $\widetilde{E}_t(c)$ on each class $c$ at the $t$-th epoch to reveal the class-specific learning status.
\begin{equation}\label{eq:4}
\widetilde{E}_t(c) = \left\{
	\begin{array}{ll}
    \frac{1}{C}, &  t = 0\\
    m \widetilde{E}_{t-1}(c) + (1-m) \frac{1}{N} \sum_{i=1}^N p^c(x_i, \theta), & t > 0
    \end{array}.
\right.
\end{equation}
Accordingly, we obtain local threshold $\widetilde{T}_t(c)$ for each class $c$ by normalizing $\widetilde{E}_t(c)$ and integrating it with global threshold $T_t$ as:
\begin{equation}\label{eq:5}
    \widetilde{T}_t(c)= \frac{\widetilde{E}_t(c)}{max\{\widetilde{E}_t(c:c \in [C]\}} T_t.
\end{equation}

On the one hand, the design of our global threshold ensures that sufficient clean samples are identified and learned by the network.
On the other hand, the design of our local threshold ensures that selected clean samples are class-balanced.
Finally, by unifying our proposed global and local thresholds, we divide the training set $D_{train}$ into a clean subset $D_{c}$ and a noisy subset $D_{n}$ in each epoch according to Eq.~\eqref{eq:6_7}.
\begin{equation}\label{eq:6_7}
\left\{
    \begin{array}{l}
    \begin{aligned}
        D_{c} &= \{(x_i, y_i)| (x_i, y_i) \in D_{train}, p^{y_i}(x_i, \theta) > \widetilde{T}_t(y_i)\} \\
        D_{n} &= \{(x_i, y_i)| (x_i, y_i) \in D_{train}, (x_i, y_i) \not \in D_{c} \} \\
    \end{aligned}
    \end{array}.
\right.
\end{equation}

\subsection{Adaptive and Balanced Re-weighting}\label{SCR}
Recent researches propose to cope with noisy samples in a semi-supervised-learning-like (SSL-like) manner by integrating sample selection and label correction \cite{dividemix,josrc}.
Identified clean samples are used conventionally for model training, while detected noisy samples are assigned pseudo labels to correct their supervision before being used for training.
However, existing methods tend to treat label-corrected noisy samples equally, neglecting their difference in reliability.
Moreover, due to different learning difficulties in various categories, label correction results may be imbalanced (noisy samples are more likely to be assigned labels of simple classes), resulting in biased label correction and sub-optimal model performance.

To mitigate the above issue, we propose a self-adaptive and class-balanced re-weighting (SCR) mechanism to adaptively assign different weights to samples according to their confidence.
Specifically, we use a temporally averaged model (\ie, mean-teacher model $\theta^*$) to generate reliable pseudo labels for detected noisy samples. 
By introducing the historical models, we obtain corrected labels $y^{corr}$ using $\theta^*$ to promote the reliability of label correction and alleviate error-propagation issues.
The mean-teacher model $\theta^*$ is not updated in the gradient back-propagation.
$\theta^*$ is updated in each training step $t^\prime$ as follows:
\begin{equation}\label{eq:8}
\theta_{t^\prime}^* = \alpha \theta_{t^\prime-1}^* +(1-\alpha) \theta_{t^\prime},
\end{equation}
in which $\theta_0^*$ is initialized using the initial model parameters of $\theta$.
Accordingly, noisy samples are assigned pseudo labels as follows:
\begin{equation}\label{eq:9}
y_i^{corr} =  \mathop{\arg \max}\limits_{j = 1,...,C} p^j(x_i, \theta^*).
\end{equation}

As mentioned above, the label correction results could be imbalanced due to the biased capability of the network.
Consequently, we propose a re-weighting method to adaptively assign larger weights to (noisy) samples with higher correction confidence.
We employ the prediction probability \wrt the corrected label to reveal the correction confidence.
Inspired by the semi-supervised learning methods \cite{MarginMatch, chen2023softmatch, mixmatch, FullMatch}, we propose to fit the underlying sample weights to a dynamic truncated normal distribution, whose mean and variance values at the $t$-th epoch are $\mu_t$ and $\sigma_t$.
The sample weights are therefore derived in a self-adaptive fashion as:
\begin{equation}\label{eq:10}
\lambda(x_i) = \left\{
	\begin{array}{ll}
    \lambda_{m} exp(\frac{(p^{y_i^{corr}}(x_i, \theta)-\mu_t)^2}{-2 \sigma_t^2}), &  p^{y_i^{corr}}(x_i, \theta) < \mu_t \\
    \lambda_{m},& otherwise
    \end{array},
\right.
\end{equation}
in which $\lambda_{m}$ is the upper bound of sample weights. 
Assuming sample weights to follow the dynamic truncated normal distribution is equivalent to treating the deviation of correction confidence from $\mu_t$ as a proxy measure of the correctness of the label correction.
Samples with higher confidence are less prone to be erroneously label-corrected than those with lower confidence, thus being assigned larger weights.

Moreover, to enable class-balanced re-weighting and promote training stability, we propose to estimate $\mu_t(c)$ and $\sigma_t^2(c)$ for each class $c$ based on their historical estimations using EMA:
\begin{equation}\label{eq:11}
\mu_t(c)= \left\{
	\begin{array}{ll}
    \frac{1}{C},&  t=0\\
    m \mu_{t-1}(c) + (1 - m)\widetilde{\mu}(c) , & t>0
    \end{array},
\right.
\end{equation}
\begin{equation}\label{eq:12}
\sigma_t^2(c)= \left\{
	\begin{array}{ll}
    1.0,&  t=0\\
    m \sigma_{t-1}^2(c) + (1 - m)\widetilde{\sigma}^2(c) , & t>0
    \end{array},
\right.
\end{equation}
in which,
\begin{equation}\label{eq:13}
\widetilde{\mu}(c) = \frac{1}{|D_{n}|} \sum_{i=1}^{|D_{n}|} p^{y_i^{corr}}(x_i, \theta),\ \  if \ y_i^{corr}=c,
\end{equation}
\begin{equation}\label{eq:14}
\widetilde{\sigma}^2(c)  = \frac{1}{|D_{n}|} \sum_{i=1}^{|D_{n}|} (p^{y_i^{corr}}(x_i, \theta)-\widetilde{\mu}(c) )^2,\ \ if \  y_i^{corr}=c.
\end{equation}

$\mu_t$ and $\sigma_t$ of the dynamic truncated normal distribution can be adaptively estimated from the correction confidence distribution based on Eqs.~\eqref{eq:11} and \eqref{eq:12}. 
As the model performance improves during training,  $\mu_t$ gradually increases and $\sigma_t$ decreases.
Since the tail of the normal distribution grows exponentially tighter, the samples with lower correction confidence are given lower weights.
Besides, we estimate class-specific $\mu_t$ and $\sigma_t$. 
This effectively alleviates the class imbalance in the label correction process caused by the biased model ability.

\begin{algorithm}[t]
\caption{Our proposed SED algorithm}
\begin{flushleft}
\vspace{-0.3cm}
\textbf{Input:} The training set $D_{train}$, network $\theta$, mean-teacher network $\theta^*$, total epochs $E_{total}$, batch size $bs$.
\vspace{-0.3cm}
\end{flushleft}
\begin{algorithmic}[1] 
    \FOR {$epoch=1,2,\ldots,E_{total}$}
    \STATE Obtain $T_t$ and $\widetilde{T}_t$ by Eqs. (\ref{eq:3}), (\ref{eq:4}) and (\ref{eq:5}) 
    \STATE Obtain $D_{c}$ and $D_{n}$ based on Eq. (\ref{eq:6_7}).
    \STATE Obtain $y^{corr}$, $\widetilde{\mu}$, and $\widetilde{\sigma}^2$ by Eqs. (\ref{eq:9}), (\ref{eq:11}) and (\ref{eq:12}) .
    \STATE Obtain $\lambda(x)$ based on Eq. (\ref{eq:10}).
    \FOR {$iteration=1,2,\ldots$}
    \STATE Fetch $B=\{(x_i, y_i)\}^{bs}$ from $D_{train}$
    \STATE Obtain ${{B}_{clean}} \subseteq D_{c}$ and ${{B}_{noise}}\subseteq D_{n}$
    \STATE Calculate $\mathcal{L} = \mathcal{L}_{D_{c}} + \mathcal{L}_{D_{n}} + \mathcal{L}_{reg}$
    \STATE Update $\theta$ by optimizing $\mathcal{L}$
    \STATE Update $\theta^*$ by Eq. (\ref{eq:8})
    \ENDFOR
    \ENDFOR
\end{algorithmic}
\begin{flushleft}
\vspace{-0.3cm}
\textbf{Output:} Updated network $\theta$.
\vspace{-0.3cm}
\end{flushleft}
\label{alg}
\end{algorithm}
\subsection{Overall Framework}
In summary, our proposed SED follows the paradigm that integrates sample selection, label correction, and sample re-weighting for addressing noisy labels. 
Details of our SED are shown in Fig.~\ref{figure2} and Algorithm {\color{red}1}.

Firstly, SED divides $D_{train}$ into a clean subset $D_{c}$ and a noisy subset $D_{n}$ in a self-adaptive and class-balanced manner. 
For samples in the clean subset $D_{c}$, we take their given labels to calculate the classification loss $\mathcal{L}_{D_{c}}$ as follow:
\begin{equation}
\label{eq:16}
\mathcal{L}_{D_{c}} = -\frac{1}{|{D_{c}}|}
\sum_{({x},{y})\in {D_{c}}}{y}\ log\ p({x},\theta).
\end{equation}
For samples in the noisy subset $D_{n}$, we discard their given labels and perform label correction based on a mean-teacher model using Eq.~\eqref{eq:9}.
Then, we calculate the loss of the noisy subset $\mathcal{L}_{D_{n}}$ as 
\begin{equation}
\label{eq:17}
\mathcal{L}_{D_{n}} = -\frac{1}{|{D_{n}}|}
\sum_{({x},{y})\in {D_{n}}} \lambda(x) {y^{corr}}\ log\ p({\hat{x}},\theta),
\end{equation}
in which $\hat{x}$ denotes the strongly-augmented view of the sample $x$.
$\lambda(x)$ represents the sample weight computed by Eq.~\eqref{eq:10}.
Finally, we incorporate an additional weighted classification loss on clean samples \wrt corrected labels (similar to $\mathcal{L}_{D_{n}}$) to further enhance the robustness of the model.
This loss term implicitly encourages prediction consistency between weakly- and strongly-augmented views of samples from the clean subset, regularizing the model to achieve better performance. 
Thus, we term this loss as the consistency regularization loss and compute it as follows:
\begin{equation}
\label{eq:18}
\mathcal{L}_{reg} = -\frac{1}{|{D_{c}}|}
\sum_{({x},{y})\in {D_{c}}}\lambda(x) {y^{corr}}\ log\ p({\hat{x}},\theta),
\end{equation}
where $\lambda(x)$ is also computed based on Eq.~\eqref{eq:10}.
Accordingly, the final objective loss function in our SED is:
\begin{equation}
\label{eq:15}
\mathcal{L} = \mathcal{L}_{D_{c}} + \mathcal{L}_{D_{n}} + \mathcal{L}_{reg}.
\end{equation}

\section{Experiments}
In this section, we conduct experiments on two synthetically corrupted datasets (\ie, CIFAR100N and CIFAR80N \cite{josrc}) and three real-world datasets (\ie, Web-Aircraft, Web-Car, and Web-Bird \cite{sun2020crssc}). 
We demonstrate the superiority of our method in coping with noisy labels by comparing SED with various state-of-the-art (SOTA) methods.
Moreover, we conduct extensive ablation studies to evaluate the effectiveness of each component in our SED.

\begin{table*}[t]
\renewcommand\tabcolsep{5pt}
\centering
\caption{Average test accuracy (\%) on CIFAR100N and CIFAR80N over the last ten epochs. Experiments are conducted under various noise conditions (“Sym” and “Asym” denote the symmetric and asymmetric label noise, respectively). Results of existing methods are mainly drawn from \cite{Co-LDL}. $^{\dagger}$ means that we re-implement the method using its open-sourced code and default hyper-parameters.}
\resizebox{1.0\linewidth}{!}{
\begin{tabular}{rccccccccccc}
\toprule
\multirow{2}{*}{\textbf{Methods}}  & \multirow{2}{*}{\textbf{Publication}}  & 
\multicolumn{3}{c}{\textbf{CIFAR100N}} & \multicolumn{3}{c}{\textbf{CIFAR80N}} \\  \cmidrule(r){3-5} \cmidrule(r){6-8}
& & Sym-20\% 	& Sym-80\%    & Asym-40\%  	& Sym-20\% 	& Sym-80\%    & Asym-40\%  \\ 
\bottomrule
Standard & - & 35.14  & 4.41 & 27.29 & 29.37 & 4.20 & 22.25 \\
Decoupling \cite{Decoupling} & NeurIPS 2017 & 33.10  & 3.89 & 26.11 & 43.49 & 10.1 & 33.74 \\
Co-teaching \cite{co-teaching} & NeurIPS 2018 & 43.73 & 15.15 & 28.35 & 60.38  & 16.59 & 42.42 \\
Co-teaching+ \cite{Co-teaching+} & ICML 2019 & 49.27  & 13.44 & 33.62 & 53.97 & 12.29 & 43.01 \\
JoCoR \cite{JoCoR} & CVPR 2020 & 53.01  & 15.49 & 32.70 & 59.99 & 12.85 & 39.37 \\
Jo-SRC \cite{josrc} & CVPR 2021 & 58.15  & 23.80 & 38.52 & {65.83} & 29.76 & 53.03 \\
SELC \cite{SELC} & IJCAI 2022 & 55.44  & 23.54 & 45.19 & 57.51 & 22.79 & 47.50 \\
DivideMix \cite{dividemix} & ICLR 2020 &57.76&28.98&43.75& 57.47 &21.18 &37.47 \\
Co-LDL \cite{Co-LDL} & TMM 2022 & 59.73  &25.12  & {52.28}  & 58.81 & 24.22 & 50.69 \\
UNICON$^{\dagger}$ \cite{UNICON} & CVPR 2022 & 55.10  &31.49  & 49.90  & 54.50 & 36.75 & 51.50 \\
NCE$^{\dagger}$ \cite{NCE} & ECCV 2022 & 54.58  &{35.23}  & 49.90  & 58.53 & {39.34} & 56.40 \\
SOP$^{\dagger}$ \cite{SOP} & ICML 2022 & 58.63  &{34.23}  & 49.87  & 60.17 & 34.05 & {53.34} \\
SPRL$^{\dagger}$ \cite{SPRL} &PR 2023 & 57.04  &28.61  & 49.38  & 47.90 & 22.25 & 40.86 \\
AGCE$^{\dagger}$ \cite{AGCE} &TPAMI 2023 & 59.38  &27.41  & 43.04  & 60.24 & 25.39 & 44.06 \\
DISC$^{\dagger}$ \cite{DISC} &CVPR 2023 & {60.28}  &33.90  & 50.56  & 50.33 & {38.23} & 47.63 \\
\bottomrule
\rowcolor{gray!20}\textbf{Ours} & - & \textbf{66.50} & \textbf{38.15} &	\textbf{58.29} & \textbf{69.10} & \textbf{42.57} & \textbf{60.87} \\
\toprule
\end{tabular}}
\label{table:1}
\vspace{-0.6cm}
\end{table*}

\subsection{Experiment Setup}
\textbf{Synthetically Corrupted Datasets.}
CIFAR100N and CIFAR80N are mainly derived from CIFAR100 \cite{CIFAR}. 
CIFAR100 consists of 60,000 RGB images (50,000 for training and 10,000 for testing).
We follow \cite{josrc} to create the closed-set noisy dataset CIFAR100N and the open-set noisy dataset CIFAR80N.
In particular, to construct the open-set noisy dataset CIFAR80N, we regard the last 20 categories in CIFAR100 as out-of-distribution samples. 
We adopt two classical noise structures: symmetric and asymmetric, with a noise ratio $n \in (0,1)$.

\noindent
\textbf{Real-World Datasets.}
To further verify the effectiveness of our SED in practical scenarios, we conduct experiments on the three real-world noisy datasets (\ie, Web-Aircraft, Web-Car, and Web-Bird \cite{sun2020crssc}), whose training images are crawled from web image search engines. 
The noise rates and structures of real-world datasets are all unknown.
No label verification information is provided.

\noindent
\textbf{Implementation Details.}
On synthetically corrupted datasets, we follow \cite{josrc} to conduct experiments with a seven-layer CNN network as the backbone.
The network is trained using SGD with a momentum of 0.9 for 100 epochs (including 20 warm-up epochs).
The batch size is 128, and the initial learning rate is 0.05.
For real-world datasets, we follow \cite{Co-LDL}  and leverage ResNet50 \cite{Resnet} pre-trained on ImageNet as our backbone.
We use the SGD optimizer with a momentum of 0.9 to train the network for 110 epochs.
The batch size, the initial learning rate, and the weight decay are 32, 0.005, and 0.0005.
The learning rate decays in a cosine annealing manner.
We train the network for 110 epochs, in which the first 10 epochs are warm-up.
The EMA coefficients $m$ and $\alpha$ are set to 0.99 and 0.95.
$\lambda_m$ is set to 1.0 for all datasets.

\noindent
\textbf{Baselines.}
For CIFAR100N and CIFAR80N, we compare our method with the following SOTA methods: Decoupling \cite{Decoupling}, Co-teaching \cite{co-teaching}, Co-teaching+ \cite{Co-teaching+}, JoCoR \cite{JoCoR}, Jo-SRC \cite{josrc}, SELC \cite{SELC}, Co-LDL \cite{Co-LDL}, UNICON \cite{UNICON}, SOP \cite{SOP}, AGCE \cite{AGCE}, and DISC \cite{DISC}.
For Web-Aircraft, Web-Bird, and Web-Car, besides the above methods, we additionally compare SED with other SOTA methods (\eg, PENCIL \cite{PENCIL}, Hendrycks \al \cite{ss-ood}, mCT-S2R \cite{mCT-S2R}, AFM \cite{AFM}, and Self-adaptive \cite{Self-adaptive}).
Moreover, we perform conventional training using the entire noisy dataset. The result is provided as a baseline (denoted as Standard).
Results in Tables \ref{table:1} and \ref{table:2} are mainly obtained from \cite{josrc} and \cite{Co-LDL}.

\begin{figure*}[t]
\centering
\includegraphics[width=\linewidth]{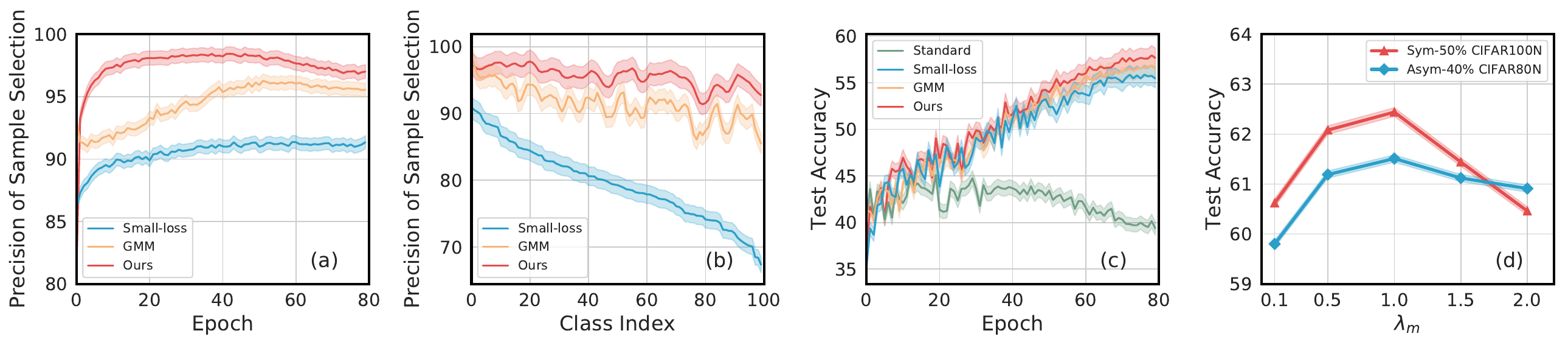}
\caption{Comparison of different sample selection methods and the ablation results of the parameter $\lambda_m$. (a) The overall precision of sample selection (\%) \vs epochs. (b) The class-wise precision of sample selection (\%) \vs classes. (c) The test accuracy (\%) \vs epochs. (d) The test accuracy (\%) of using different $\lambda_m$. }
\label{figure5}
\vspace{-0.2cm}
\end{figure*}

\subsection{Evaluation on Synthetic Datasets}
We show the comparison results between our SED and existing SOTA methods on the synthetic datasets (\ie, CIFAR100N and CIFAR80N) in Table~\ref{table:1}.

\noindent
\textbf{Results on CIFAR100N.}
Table~\ref{table:1} shows that SED consistently achieves the best performance compared to SOTA methods on CIFAR100N.
In particular, it should be noted that SED can better adapt to severely noisy situations (\ie, Sym-80\%), while most SOTA approaches almost fail in the most inferior case.
It should be emphasized that the asymmetric noise case is often more challenging than the symmetric one.
Our SED shows a significant improvement (\ie, $\ge 6.01\%$) on Asym-40\%.
Experiments on CIFAR100N show that SED can effectively deal with closed-set label noise in different noise situations.

\noindent
\textbf{Results on CIFAR80N.}
To simulate real-world scenarios, CIFAR80N contains both closed-set and open-set noisy labels, making it undoubtedly more challenging.
Results shown in Table~\ref{table:1} illustrate:
(1) in the case of Sym-20\%, our SED can achieve a 3.27\% performance improvement.
(2) in the case of Sym-80\%, while most SOTA approaches fail to tackle the massive noisy labels, SED achieves the best result.
(3) when the noise scenario becomes harder (\ie, Asym-40\%), our SED consistently obtains the best performance, outperforming the second-best result by 7.53\%.
Table~\ref{table:1} proves that SED performs consistently better than existing methods when coping with open-set noisy datasets.

\begin{table*}[t]
\renewcommand\tabcolsep{5pt}
\centering

\caption{The comparison with SOTA approaches in test accuracy (\%) on real-world noisy datasets: Web-Aircraft, Web-Bird, Web-Car. Results of existing methods are mainly drawn from \cite{Co-LDL}. $^{\dagger}$ means that we re-implement the method using its open-sourced code and default hyper-parameters.}
\label{table:2}
\resizebox{1.0\linewidth}{!}{
\begin{tabular}{rcccccc}
\toprule
\multirow{2}{*}{\textbf{Methods}}  & \multirow{2}{*}{\textbf{Publication}} & \multirow{2}{*}{\textbf{Backbone}}  & 
\multicolumn{4}{c}{\textbf{Performances(\%)}} \\  \cline{4-7} 
& & & Web-Aircraft 	& Web-Bird    & Web-Car  & Average \\ 
\bottomrule
Standard & - & ResNet50 & 60.80  & 64.40 & 60.60 &61.93\\
Decoupling \cite{Decoupling} & NeurIPS 2017 & ResNet50 & 75.91 & 71.61 & 79.41 & 75.64\\
Co-teaching \cite{co-teaching} & NeurIPS 2018 & ResNet50 & 79.54 & 76.68 & 84.95 & 80.39\\
Co-teaching+ \cite{Co-teaching+} & ICML 2019 & ResNet50 & 74.80 & 70.12 & 76.77 & 73.90\\
PENCIL \cite{PENCIL} & CVPR 2019 & ResNet50 & 78.82 & 75.09 & 81.68 & 78.53\\
Hendrycks \al \cite{ss-ood} & NeurIPS 2019 & ResNet50 & 73.24 & 70.03 & 73.81 & 72.36\\
mCT-S2R \cite{mCT-S2R} & WACV 2020 & ResNet50 & 79.33 & 77.67 & 82.92 & 79.97\\
JoCoR \cite{JoCoR} & CVPR 2020 & ResNet50 & 80.11 & 79.19 & 85.10 & 81.47\\
AFM \cite{AFM} & ECCV 2020 & ResNet50 & 81.04 & 76.35 & 83.48 & 80.29\\
DivideMix \cite{dividemix} & ICLR 2020 & ResNet50 & 82.48 & 74.40 & 84.27 & 80.38 \\
Self-adaptive \cite{Self-adaptive} & NeurIPS 2020 & ResNet50 & 77.92 & 78.49 & 78.19 & 78.20\\
Co-LDL \cite{Co-LDL} & TMM 2022 & ResNet50 & 81.97 & 80.11 & 86.95 & 83.01\\
UNICON$^{\dagger}$ \cite{UNICON} & CVPR 2022 & ResNet50 & 85.18 & {81.20} & 88.15 & 84.84\\
NCE $^{\dagger}$ \cite{NCE} & ECCV 2022 & ResNet50 & 84.94 & 80.22 & 86.38 & 83.85 \\
SOP$^{\dagger}$ \cite{SOP} & ICML 2022 & ResNet50 & 84.06 & 79.40 & 85.71 & 83.06 \\
SPRL$^{\dagger}$ \cite{SPRL} & PR 2023 & ResNet50 & 84.40 & 76.36 & 86.84 & 82.53 \\
AGCE$^{\dagger}$ \cite{AGCE} & TPAMI 2023 & ResNet50 & 84.22 & 75.60 & 85.16  & 81.66 \\
DISC$^{\dagger}$ \cite{DISC} & CVPR 2023 & ResNet50 &  {85.27} & 81.08 & {88.31} & {84.89} \\
\bottomrule
\rowcolor{gray!21}\textbf{Ours}  & - & ResNet50 & \textbf{86.62} & \textbf{82.00} &	\textbf{88.88} &\textbf{85.83} \\
\toprule
        \end{tabular}}
        \vspace{-0.2cm}
\end{table*}
\subsection{Evaluation on Real-world Datasets}
Table~\ref{table:2} shows the experimental results of existing methods and SED on Web-Aircraft, Web-Bird, and Web-Car, which contain open-set and closed-set noise simultaneously.
From this table, we can find that SED can achieve better (or comparable) performance against SOTA approaches in different datasets.
SED achieves performances of 86.62\%, 82.00\%, and 88.88\% on test sets of Web-Aircraft, Web-Bird, and Web-Car, respectively.
The average test accuracy outperforms existing SOTA methods by 0.94\%.
It should be noted that the second and third-best methods (\ie, DISC and UNICON) involve the Mixup training trick and two simultaneously trained networks respectively, while SED trains only one network without Mixup.
Compared to existing methods, our SED eliminates the demand for dataset-dependent prior knowledge (\eg, pre-defined drop rate/threshold), making it easier to adapt to different datasets.

\subsection{Ablation Studies}
In this section, we demonstrate the effectiveness of each component in our SED (\ie, SCS, SCR, and CR).
Besides, we investigate the effect of the hyper-parameter $\lambda_m$ in Eq.~\eqref{eq:10}.
Unless otherwise stated, ablation experiments are conducted on CIFAR100N (Sym-50\%).
Table ~\ref{table:3} and Table~\ref{table:4} show the impact of each component. 

\noindent
\textbf{Effects of Self-adaptive and Class-balanced Sample Selection.}
As analyzed above, existing sample selection methods tend to struggle with the demand for dataset-dependent prior knowledge, such as pre-defined drop rate/threshold.
However, these hyper-parameters are usually unknown and hard to estimate in real-world datasets.
The proposed SCS strategy in our method allows adaptive sample selection in a class-balanced manner, making our SED have better generalization performance on different datasets.
As shown in Table~\ref{table:3}, employing SCS achieves a 24.11\% performance gain compared to the baseline Standard.
We also provide the result of using SCS without local thresholds and global thresholds.
This proves that our threshold design is crucial for improving the robustness of the model.

To further demonstrate the superiority of our SCS over previous sample selection strategies, we compare our SCS with two commonly-used methods (\ie, small-loss \cite{co-teaching}, and GMM \cite{dividemix}) in Fig.~\ref{figure5}.
As shown in Fig.~\ref{figure5} (a), our SCS is shown to be more effective in selecting clean samples accurately compared with the other two strategies.
Additionally, we compare the sample selection accuracy for each category in the selected clean subset and present the comparison in Fig.~\ref{figure5} (b).
It illustrates that the selection results of SCS are more balanced.
The curves of test accuracy are shown in Fig.~\ref{figure5} (c), revealing the leading performance of our SCS compared with the other two methods and the baseline.

\begin{table}[t]
\renewcommand\tabcolsep{15pt}
\centering
\caption{Effect of each component in the test accuracy (\%) on CIFAR100N.}
\resizebox{0.75\linewidth}{!}{
\begin{tabular}{l|c}
\toprule
\textbf{Model} & \textbf{Test Accuracy}\\
\midrule
Standard & 34.10\\
\midrule
Standard+SCS w/o local threshold & 53.36\\
Standard+SCS w/o global threshold & 55.64\\
Standard+SCS w/o EMA & 54.72\\
Standard+SCS & 58.21\\
Standard+SCS+SCR w/o re-weighting & 59.75\\
Standard+SCS+SCR w/o EMA & 60.08\\
Standard+SCS+SCR & 60.43\\
Standard+SCS+SCR+CR & 62.65\\
\bottomrule
        \end{tabular}}
\label{table:3}
\vspace{-0.4cm}
\end{table}
\noindent
\textbf{Effects of Self-adaptive and Class-balanced Sample Re-weighting.}
Our SED follows an SSL-like paradigm. Selected clean samples are learned conventionally, while detected noisy samples are also fed into the network for training after label correction.
However, the biased model capability tends to result in imbalanced label correction, hurting the model performance. 
We accordingly propose SCR to re-weight detected noisy samples in a self-adaptive and class-balanced manner when using their corrected labels for training.
Table~\ref{table:3} shows a performance gain of 2.19\% by employing our proposed SCR.
The only involved hyper-parameter in the SCR is the $\lambda_m$ in Eq.~\eqref{eq:10}.
Fig.~\ref{figure5} (d) exhibits the influence of different $\lambda_m$ values on the test accuracy when experimenting with CIFAR100N (Sym-50\%) and CIFAR80N (Asym-40\%).
It can be observed that the best performance is achieved when $\lambda_m = 1.0$ on CIFAR100N (Sym-50\%) and CIFAR80N (Asym-40\%).

\begin{table}[t]
\centering

\renewcommand\tabcolsep{10pt}
    \centering
        \caption{Effect of promoting class balance on CIFAR100N (left) and CIFAR80N (right). Test accuracy (\%) of SED with and without the class-balanced design is compared under different settings.} \label{table:4}
    \begin{subtable}{0.45\textwidth}
        
\resizebox{1.0\linewidth}{!}{
         \begin{tabular}{c|c|c}
         \toprule
        Class-balanced? & \XSolidBrush  & \Checkmark\\
         \hline
        Sym-20\% &64.16&66.59	\\
        \hline
        Sym-80\% &38.08 &39.32\\
        \hline
        Asym-40\% &52.78 &58.80 \\ 
         \bottomrule
        \end{tabular}}
        \end{subtable}
        \hspace{0.2cm}
        \begin{subtable}{0.45\textwidth}
        \centering
\resizebox{\linewidth}{!}{
         \begin{tabular}{c|c|c}
         \toprule
        Class-balanced? & \XSolidBrush  & \Checkmark\\
         \hline
        Sym-20\% &67.20&68.75\\
        \hline
        Sym-80\%&39.74&42.90\\
        \hline
        Asym-40\%  &57.00&61.51 \\ 
         \bottomrule
        \end{tabular}}

    \end{subtable}
\end{table}

\noindent
\textbf{Effects of Consistency Regularization.}
Although clean samples selected by SED are more accurate and balanced than previous methods, it is inevitable that some noisy data will be mistakenly selected into the clean subset.
Therefore, we impose an additional CR on the selected clean samples to enhance the model's robustness.
Table~\ref{table:3} shows that CR successfully boosts model performance by 2.02\%, revealing the benefits that CR brings to our model.

\noindent
\textbf{Effects of Promoting Class Balance.}
As stated in SCS and SCR, our SED favors the class-balanced design.
Specifically, SCS estimates local thresholds on each class to avoid imbalanced sample selection, while SCR also estimates $\mu_t$ and $\sigma_t^2$ of the dynamic truncated normal distribution for each class to encourage balanced re-weighting.
As shown in Table~\ref{table:4}, we investigate the effect of the class-balanced design in SED.
We can find that our method consistently achieves better performance when incorporated with the class-balanced design, especially in harder scenarios.
Table~\ref{table:4} effectively demonstrates that the class-balanced design in our SED is beneficial for model performance.

\section{Conclusion}
In this paper, we proposed a simple yet effective approach named SED to address the inferior model performance caused by noisy labels.
We designed a self-adaptive and class-balanced sample selection strategy to distinguish between clean and noisy samples.
Clean samples were learned conventionally. A mean-teacher model was employed to correct the labels of detected noisy samples.
Subsequently, SED re-weighted noisy samples in a self-adaptive and class-balanced fashion based on the correction confidence when leveraging them for model training.
Finally, we additionally imposed consistency regularization on the clean subset to further improve model performance.
Comprehensive experiments and ablation analysis on synthetic and real-world noisy datasets validated the superiority of our SED.

\bibliographystyle{splncs04}
\bibliography{egbib}
\end{document}